\documentclass{article}

\usepackage[nonatbib,preprint]{neurips_2019}
\usepackage[numbers,sort]{natbib}

\usepackage[utf8]{inputenc} 
\usepackage[T1]{fontenc}    
\usepackage{hyperref}       
\usepackage{url}            
\usepackage{booktabs}       
\usepackage{amsfonts}       
\usepackage{nicefrac}       
\usepackage{microtype}      
\usepackage{amsmath}
\usepackage{comment}
\usepackage{graphicx}
\usepackage{subcaption}
\usepackage{algorithm}
\usepackage{algorithmic}
\usepackage{wrapfig}
\usepackage{placeins}

\setcitestyle{square}

\newcommand{\bs}{\mathbf{s}}
\newcommand{\ba}{\mathbf{a}}
\newcommand{\states}{\mathcal{S}}
\newcommand{\actions}{\mathcal{A}}

\title{Reward-Conditioned Policies}

\author{%
  Aviral Kumar, Xue Bin Peng, Sergey Levine \\
  Department of Electrical Engineering and Computer Sciences\\
  University of California, Berkeley\\
  \texttt{aviralk@berkeley.edu}, \texttt{xbpeng@berkeley.edu}, \texttt{svlevine@eecs.berkeley.edu} \\
}

\begin{document}

\maketitle

\begin{abstract}
Reinforcement learning offers the promise of automating the acquisition of complex behavioral skills. However, compared to commonly used and well-understood supervised learning methods, reinforcement learning algorithms can be brittle, difficult to use and tune, and sensitive to seemingly innocuous implementation decisions. In contrast, imitation learning utilizes standard and well-understood supervised learning methods, but requires near-optimal expert data. Can we learn effective policies via supervised learning without demonstrations? The main idea that we explore in this work is that non-expert trajectories collected from sub-optimal policies can be viewed as optimal supervision, not for maximizing the reward, but for matching the reward of the given trajectory. By then conditioning the policy on the numerical value of the reward, we can obtain a policy that generalizes to larger returns. We show how such an approach can be derived as a principled method for policy search, discuss several variants, and compare the method experimentally to a variety of current reinforcement learning methods on standard benchmarks.
\end{abstract}

\section{Introduction}

Reinforcement learning, particularly when combined with high-capacity function approximators such as deep networks, has the potential to automatically acquire control strategies for complex tasks together with the perception and state estimation machinery needed to accomplish them, all the while requiring minimal manual engineering~\cite{singh2019viceraq,kalashnikov18qtopt}. However, in practice, such reinforcement learning methods suffer from a number of major drawbacks that have limited their utility for real-world problems. Current deep reinforcement learning methods are notoriously unstable and sensitive to hyperparameters~\cite{fu19diagnosing,henderson2017reinforcement}, and often require a very large number of samples. In this paper, we study a new class of reinforcement learning methods that allow simple and scalable supervised learning techniques to be applied directly to the reinforcement learning problem.

A central challenge with adapting supervised learning methods to autonomously learn skills defined by a reward function is the lack of optimal supervision: in order to learn behaviors via conventional supervised learning methods, the learner must have access to labels that indicate the optimal action to take in each state. The main observation in our work is that \emph{any} experience collected by an agent can be used as optimal supervision \emph{when conditioned on the quality of a policy}. That is, actions that lead to mediocre returns represent ``optimal'' supervision \emph{for a mediocre policy}. We can implement this idea in a practical algorithm by learning policies that are conditioned on the reward that will result from running that policy, or other quantities derived from the reward, such as the advantage value. In this way, all data gathered by the agent can be used as ``optimal'' supervision for a particular value of the conditioning return or advantage.

Building on this insight, we propose to learn policies of the form $\pi_\theta(\mathbf{a}|\mathbf{s},Z)$, where $\theta$ represents the parameters of the policy, $\mathbf{a}$ represents the action, $\mathbf{s}$ represents the state, and $Z$ represents some measure of value -- either the total return, or the advantage value of $\ba$ in state $\bs$. Any data collected using \emph{any} policy can provide optimal supervision tuples of the form $(\mathbf{s},Z,\mathbf{a})$, and a policy of this form can be trained on such data using standard supervised learning.

Our main contribution is a practical reinforcement learning algorithm that uses standard supervised learning as an inner-loop sub-routine. We show how reward-conditioned policies can be derived in a principled way from a policy improvement objective, discuss several important implementation choices for this method, and evaluate it experimentally on standard benchmark tasks and fully off-policy reinforcement learning problems. We show that some variants of this method can perform well in practice, though a significant gap still exists between this approach and state-of-the-art reinforcement learning algorithms.

\section{Related Work}
\label{sec:related}

Most current reinforcement learning algorithms aim to either explicitly compute a \emph{policy gradient}~\citep{Williams1992,PoliGrad1999}, accurately fit a value function or Q-function~\citep{Watkins92q,Precup2001,mnih2015humanlevel,NAF16}, or both~\citep{DDPG2016,haarnoja18b}. While such methods have attained impressive results on a range of challenging tasks~\citep{mnih2015humanlevel,Levine2016,HeessTSLMWTEWER17,2018-TOG-deepMimic,Rajeswaran-RSS-18}, they are also known to be notoriously challenging to use effectively, due to sensitivity to hyperparameters, high sample complexity, and a range of important and delicate implementation choices that have a large effect on performance~\citep{Hasselt2016,WangBHMMKF16,Munos2016,RainbowDQN2017,fujimoto18a,nachum2018learning,fu19diagnosing}.

In contrast, supervised learning is comparatively well understood, and even imitation learning methods can often provide a much simpler approach to learning effective policies when demonstration data is available~\citep{alvinn, imitation_endtoend,imitation_tutorial}. Indeed, a number of recent works have sought to combine imitation learning and reinforcement learning~\citep{MARWIL,sun2018truncated,combining_imitation_and_rl}.
However, when expert demonstrations are not available, supervised learning cannot be used directly. A number of prior works have sought to nonetheless utilize supervised learning in the inner loop of a reinforcement learning update, either by imitating a computational expert (e.g., another more local RL algorithm)~\citep{Levine2016,DnC}, the best-performing trajectories~\citep{SIL2018}, or by reweighting sub-optimal data to make it resemble samples from a more optimal policy~\cite{Peters2010REP,Peters2007RWR}. In this paper, we utilize a simple insight to make it feasible to use suboptimal, non-expert data for supervised learning: suboptimal trajectories can be used as optimal supervision for a policy that aims to achieve a specified return or advantage value.

The central idea behind our method -- that suboptimal trajectories can serve as optimal supervision for other tasks or problems -- has recently been explored for \emph{goal}-conditioned policies, both with reinforcement learning~\cite{Kaelbling93b,Andrychowicz2017her,pong18tdm} and supervised learning~\citep{ghosh2019gcsl}. Our approach can be viewed as a generalization of this principle to arbitrary tasks, conditioning on the reward rather than a goal state. Like our method, \citet{harutyunyan2019hca} also learn the distribution of actions conditioned on future states or the trajectory return, but then utilize such models with standard RL techniques, such as policy gradients, to provide more effective credit assignment and variance reduction. Concurrently with our work, \citet{schmidhuber2019upsidedownrl} and \citet{srivastave2019upsidedownrl} proposed a closely related algorithm that also uses supervised learning and reward conditioning. While our work is concurrent, we further explore the challenges with this basic design, demonstrate that a variety of careful implementation choices are important for good performance, and provide detailed comparisons to related algorithms.

\section{Preliminaries}
In reinforcement learning, our goal is to learn a control policy that maximizes the expected long term return in a task which is modeled as a Markov decision process (MDP). At each timestep $t$, the agent receives an environment state $\bs_t \in \states$, executes an action $\ba_t \in \actions$ and observes a reward $r_t = r(\bs_t, \ba_t)$ and the next environment state $\bs_{t+1}$. The goal of the RL algorithm is to learn a policy $\pi_\theta(\ba_t | \bs_t)$ that maximizes the return, which is the cumulative discounted reward $J(\theta)$, defined as
\begin{equation}
    J(\theta) =\mathbb{E}_{\bs_0 \sim p(s_0), \ba_{0:\infty} \sim \pi, \bs_{t+1} \sim p(\cdot|\ba_{t}, \bs_{t})} \left[ \sum_{t=1}^\infty \gamma^t r(\bs_t, \ba_t) \right]. \nonumber
\end{equation}
Prior reinforcement learning methods generally either aim to compute the derivative of $J(\pi)$ with respect to the policy parameters $\theta$ directly via policy gradient methods~\cite{Williams1992}, or else estimate a value function or Q-function by means of temporal difference learning, or both. Our aim will be to avoid complex and potentially high-variance policy gradient estimators, as well as the complexity of temporal difference learning.

\section{Reward-Conditioned Policies}

\begin{algorithm}[t]
            \caption{Generic Algorithm for Reward-Conditioned Policies (RCPs)}
            \begin{algorithmic}[1]
            \STATE{$\theta_1 \leftarrow \text{random initial parameters}$}
            \STATE{$\mathcal{D} \leftarrow \emptyset$}
            \STATE{$\hat{p}_1(Z) \leftarrow $ initial value distribution}
            
            \FOR{$\text{iteration} \ k = 1, ... , k_\mathrm{max}$}
                \STATE{sample target value $\hat{Z} \sim \hat{p}_k(Z)$.}
                \STATE{roll out trajectory $\tau = \{\mathbf{s}_t, \mathbf{a}_t, r_t\}_{t=0}^T$, with policy $\pi_{\theta_k}(\cdot | \mathbf{s}_t,  \hat{Z})$}
                \STATE{for each step $t$, label $(\mathbf{s}_t,\mathbf{a}_t)$ with observed value $Z_t$}
                \STATE \text{store tuples $\{\mathbf{s}_t, \mathbf{a}_t, Z_t \}_{t=0}^T$ in $\mathcal{D}$}
                 \STATE $\theta_{k+1} \leftarrow \mathop{\mathrm{arg \ max}}_{\theta} \mathbb{E}_{\mathbf{s}, \mathbf{a}, Z \sim \mathcal{D}} \left[ \mathrm{log} \ \pi_\theta(\mathbf{a} | \mathbf{s}, Z) \right]$
                 \STATE{$\hat{p}_{k+1} \leftarrow $ update target value distribution using $\mathcal{D}$}
            \ENDFOR
            \end{algorithmic}
            \label{alg:rcp}
\end{algorithm}

The basic idea behind our approach is simple: we alternate between a training a policy of the form $\pi_\theta(\ba_t | \bs_t, Z)$ with supervised learning on all data collected so far, where $Z$ is an estimate of the return for the trajectory containing the tuple $(\bs_t, \ba_t)$, and using the latest policy to collect more data. We first provide an overview of the generic RCP algorithm, and then describe two practical instantiations of the method.

\subsection{Reward-Conditioned Policy Training}

The generic RCP algorithm is summarized in Algorithm~\ref{alg:rcp}. At the start of each rollout, a target value $\hat{Z}$ is sampled from the current target distribution $\hat{Z} \sim \hat{p}_k(Z)$. The current policy $\pi_{\theta_k}(\mathbf{a}|\mathbf{s}, \hat{Z})$ is then conditioned on $\hat{Z}$ and used to sample a trajectory $\tau_k$ from the environment. After a rollout, each timestep $t$ is relabled with a new value $Z_t$ reflecting the actual rewards observed over the course of the rollout. This value can be the observed total reward-to-go, or the estimated advantage at $(\mathbf{s}_t,\mathbf{a}_t)$. The tuples $\{\mathbf{s}_t, \mathbf{a}_t, Z_t\}$ are then added to the dataset $\mathcal{D}$, which is structured as a first-in first-out queue. 
The reward-conditioned policy is then updated via supervised regression on the data in the buffer. Finally, the target return distribution $\hat{p}(Z)$ is updated using the data in $\mathcal{D}$, and the process is repeated. RCP performs policy updates using only supervised regression, leveraging prior suboptimal trajectories as supervision.

We explore two specific choices for the form of the values $Z$: conditioning on the total \textbf{r}eturn, which we refer to as RCP-R, and conditioning on the \textbf{a}dvantage, which we refer to as RCP-A. The return conditioned variant, RCP-R, is the simplest: here, we simply choose $Z_t$ to be the discounted reward to go along the sampled trajectory, such that $Z_t = \sum_{t'=t}^T \gamma^{t'-t} r_{t'}$.

A more complex but also more effective version of the algorithm can be implemented by conditioning on the \emph{advantage} of $\ba_t$ in state $\bs_t$. The advantage function is defined as $A(\bs,\ba) = Q(\bs,\ba) - V(\bs)$, where $V(\bs)$ is the state value function, and $Q(\bs,\ba)$ is the state-action value function. Thus, RCP-A uses $Z_t = A(\bs_t,\ba_t)$, with $Q(\bs,\ba)$ estimated using a Monte Carlos estimate, and $V(\bs)$ estimated using a separately fitted value function $\hat{V}_\phi(\bs)$. Thus, we have $Z_t = \sum_{t'=t}^T \gamma^{t'-t} r_{t'} - \hat{V}_\phi(\bs_t)$. The value function can be fitted using Monte Carlo return estimates, though we opt for a TD($\lambda$) estimator, following prior work~\cite{peng19awr,Schulmanetal_ICLR2016}.

An important detail of the RCP algorithm is the update to the target value distribution $\hat{p}_k(Z)$ on line 10. We will describe the theoretical considerations for the choice of $\hat{p}_k(Z)$ in Section~\ref{sec:theory}, while here we describe the final procedure that we actually employ in our method. We represent $\hat{p}_k(Z)$ as a normal distribution, with mean $\mu_Z$ and standard deviation $\sigma_Z$. The mean and variance are updated based on the \textit{soft-maximum}, i.e. $\log\sum\exp$, of the target value $Z$ observed so far in the dataset $\mathcal{D}$.
As per line 5 in Algorithm~\ref{alg:rcp}, we sample $\hat{Z}$ from $\hat{p}_k(Z)$ for each rollout.
For RCP-A, a new sample for $Z$ is drawn at each time step, while for RCP-R, a sample for the return $Z$ is drawn once for the whole trajectory.

\subsection{Implementation and Architecture Details}
\label{sec:details}

We opt to use a deterministic policy for evaluation in accordance with the evaluation protocol commonly used in prior RL algorithms~\cite{haarnoja18b}. During evaluation, the target value is always chosen to be equal to $\mu_Z + \sigma_Z$ to avoid stochasticity arising from the target value input.

\begin{wrapfigure}{r}{0.4\textwidth} 
    \centering
    \vspace{-0.2in}
    \includegraphics[width=0.8\linewidth]{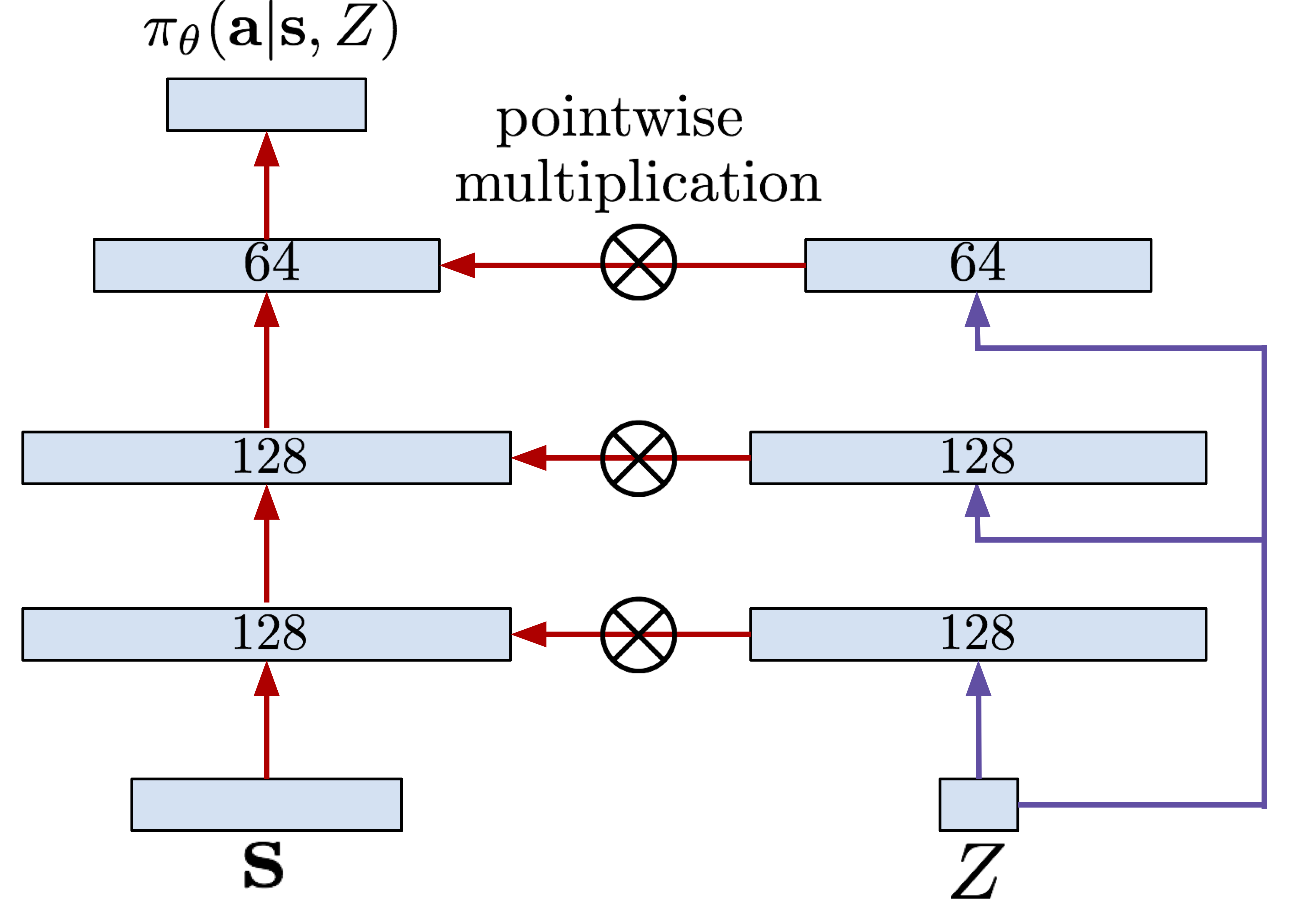}
    \caption{\footnotesize{The network architecture used for RCPs in our experiments. Inspired by \citep{cbn,film,oord16pixelcnn}, we use multiplicative interactions between an embedding of $Z$ and intermediate layers of the policy network.}}
    \vspace{-0.1in}
    \label{fig:rcp_arch}
\end{wrapfigure}
We model the policy $\pi_\theta(\ba|\bs, Z)$ as a three-layer fully-connected deep neural network that takes $s$ and $Z$ as inputs and outputs a Gaussian distribution over actions. A simple choice for the architecture of the policy network would be to concatenate the additional scalar target value $Z$ to the state $\bs$, and then use a standard multi-layer fully connected network. However, prior work has observed that such \emph{conditioning} variables can often be utilized more effectively in an architecture that incorporates multiplicative interactions~\citep{cbn,film,oord16pixelcnn}. Based on this insight, we found that using multiplicative interactions between embeddings of $Z$ and each intermediate layer of the main policy network, shown in Figure~\ref{fig:rcp_arch}, produced substantially better results in our experiments. This design prevents the policy network from ignoring the input target values.

\subsection{Theoretical Motivation for Reward-Conditioned Policies}
\label{sec:theory}

In this section, we derive the two variants of RCPs, RCP-R and RCP-A, as approximate solutions to a constrained policy search problem. This derivation resembles REPS~\cite{Peters2010REP} and AWR~\cite{peng19awr}.

\paragraph{Notation.} We denote a trajectory by $\tau$, and use $Z(\tau)$ to denote the return of the trajectory, given by $Z(\tau) = \sum_t r(\bs_t,\ba_t)$. For the purpose of this derivation, we operate in a setting where $Z(\tau)$ can be stochastic, although deterministic returns are a special case of this scenario. We refer to the joint distribution of trajectories $\tau$ and returns $Z$ as $p(\tau, Z)$. We denote the joint distribution over trajectories and returns under a sampling policy $\mu$ as $p_\mu(\tau, Z)$.

\subsubsection{Return-Conditioned Variant (RCP-R)}
\label{sec:rcp}

Our constrained policy search formulation aims to learn a return-conditioned policy $\pi_\theta(\ba|\bs, Z)$ that maximizes the discounted long-term return $J(\theta)$, under the constraint that the induced trajectory-return marginal $p_\pi(\tau, Z)$ is close to the marginal of the sampling policy, $p_\mu(\tau, Z)$. We will first compute the optimal non-parametric solution $\pi^*$ to the above described optimization problem and then learn $\pi_\theta(\ba|\bs, Z)$ by projecting $\pi^*$ into the space of parametric policies $\Pi = \{ \pi_\theta(\ba|\bs, Z) | \theta \in \Theta\}$. This can be formalized as:
\begin{align}
    \mathop{\mathrm{arg \ max}}_{\pi} \quad &
    \mathbb{E}_{\tau, {Z} \sim p_{\pi}(\tau, Z)} \left[{Z} \right] \label{eqn:rcp_objective} \\
    \textrm{s.t.} \quad & \mathrm{D_{KL}} \left( p_{\pi}(\tau, {Z}) || p_\mu(\tau, {Z}) \right) \leq \varepsilon  
    \label{eqn:RCP_update}
\end{align}
Now, we can derive the supervised regression update for RCPs as a solution to the above constrained optimization. We first form the Lagrangian of the constrained optimization problem presented above with Lagrange multiplier $\beta$:
\begin{align}
 \mathcal{L}(\pi, \beta) = \mathbb{E}_{\tau, {Z} \sim p_\pi(\tau, Z)} \left[{Z} \right] + \beta \left( \varepsilon - \mathbb{E}_{\tau, {Z} \sim \sim p_\pi(\tau, Z)} \left[ \log \frac{p_\pi(\tau, {Z})}{p_\mu(\tau, {Z})} \right] \right)
\end{align}
Differentiating $\mathcal{L}(\pi, \beta)$ with respect to $\pi$
and $\beta$ and applying optimality conditions, we obtain a non-parametric form for the joint trajectory-return distribution of the optimal policy, $p_{\pi^*}(\tau, {Z})$:
\begin{align}
    p_{\pi^*} (\tau, {Z}) \propto p_\mu(\tau, {Z}) \exp\left( \frac{{Z}}{\beta}\right)
    \label{eqn:rcpPolicy}
\end{align}
Prior work has used this derivation to motivate a \emph{weighted} supervised learning objective for the policy, where the policy is trained by regressing onto previously seen actions, with a weight corresponding to the exponentiated return $\exp({Z}/\beta)$~\cite{Peters2010REP,Peters2007RWR,peng19awr}. To instead obtain an \emph{unweighted} objective, we can instead decompose the joint distribution $p_\pi(\tau, {Z})$ into conditionals $p_\pi({Z})$ and $p_\pi(\tau|{Z})$, and use this decomposition to obtain an expression for the trajectory distribution conditioned on the target return ${Z}$. Thus, we can convert Equation~\ref{eqn:rcpPolicy} into:
\begin{align}
    p_{\pi^*} (\tau|{Z}) p_{\pi^*}({Z}) \propto \left[ {p_\mu(\tau|{Z}) p_\mu({Z})} \right] \exp\left( \frac{{Z}}{\beta}\right) 
    \label{eqn:rcp_policy_separated}
\end{align}
Equation~\ref{eqn:rcp_policy_separated} can be decomposed into separate expressions for the target distribution $p_{\pi^*}({Z})$ and the conditional trajectory distribution $p_{\pi^*}(\tau|{Z})$. We obtain a maximum likelihood objective for $p_{\pi^*}(\tau|{Z})$ and an exponentially weighted maximum-likelihood objective for the target distribution $p_{\pi^*}(\hat{Z})$. 
\begin{align}
    p_{\pi^*}(\tau|{Z}) &\propto p_\mu(\tau|{Z}) \label{eqn:rcp_policy_update}\\
    p_{\pi^*}({Z}) &\propto p_\mu({Z}) \exp\left( \frac{{Z}}{\beta} \right)
    \label{eqn:rcp_target_update}
\end{align}
Equation~\ref{eqn:rcp_policy_update} corresponds to fitting a policy $\pi^*$ to generate trajectories that achieve a particular target return value ${Z}$ as depicted in Step 9 in Algorithm~\ref{alg:rcp}. Equation~\ref{eqn:rcp_target_update} corresponds to the process of improving the expected return of a policy by updating the target return distribution to assign higher likelihoods to large values of ${Z}$ as shown in Step 10 in Algorithm~\ref{alg:rcp}.

For the final steps, we factorize $p_\pi(\tau|{Z})$ as $p_\pi(\tau|{Z}) = \Pi_{t} \pi(\ba_t|\bs_t, {Z}) p(\bs_{t+1}|\bs_t, \ba_t)$, where the product is over all time steps $t$ in a tajectory $\tau$, and the dynamics $p(\bs_{t+1}|\bs_t, \ba_t)$ are independent of the policy. To train a parametric policy $\pi_\theta(\ba|\bs, \hat{Z})$, we project the optimal non-parametric policy $p_\pi^*$ computed above onto the manifold of parametric policies, according to
\begin{align}
    \pi_\theta(\ba|\bs, {Z}) = & \mathop{\mathrm{arg \ min}}_{\theta} \quad \mathbb{E}_{{Z} \sim \mathcal{D}} \left[\mathrm{D_{KL}} \left(p_{\pi^*}(\tau| {Z}) || p_{\pi_\theta} (\tau| {Z}) \right) \right] \\
    = & \mathop{\mathrm{arg \ max}_{\theta}} \mathbb{E}_{{Z} \sim \mathcal{D}} \left[ \mathbb{E}_{\ba \sim \mu(\ba|\bs, \hat{Z})} \left[ \log \pi_\theta(\ba|\bs, {Z}) \right] \right]
    \label{eqn:RCPUpdate}
\end{align}
Equation~\ref{eqn:RCPUpdate} corresponds to a maximum likelihood update for the policy $\pi_\theta$. Training is performed only for target return values ${Z}$ that have actually been observed and are present in the buffer $\mathcal{D}$. We choose to maintain an approximate parametric Gaussian model for $p_{\pi^*}({Z})$, and continuously update this models online according to the update in Equation~\ref{eqn:rcp_target_update}. Section~\ref{sec:details} provides more details on maintaining this model in our practical implementation. 

\subsubsection{Advantage-Conditioned Variant (RCP-A)}
\label{sec:acp}

In this section, we present a derivation of the advantage-conditioned variant. Our derivation is based on the idea of learning a policy to maximize the \textit{expected improvement} over the sampling policy $\mu$. Expected improvement of policy $\pi(\ba|\bs)$ over another policy $\mu(\ba|\bs)$ is defined as the difference between their expected long-term discounted returns $\eta_\mu(\pi) = J(\pi) - J(\mu)$. Using policy difference lemma~\cite{Kakade2002}, we can express expected improvement as: 
\begin{equation}
    \eta_\mu(\pi) = J(\pi) - J(\mu) = \mathbb{E}_{\bs, \ba \sim d^\pi(\bs, \ba)} \left[A_\mu(\bs, \ba) \right] \approx \mathbb{E}_{\bs \sim d^\mu(\bs), \ba \sim \pi(\ba|\bs)}\left[A_\mu(\bs, \ba)\right] 
    \label{eqn:policy_adv_lemma}
\end{equation}
where the approximate equality holds true if $\pi$ and $\mu$ are similar~\cite{TRPOschulman15}. 

Analogous to the derivation of RCP-R, for each state-action pair $(\bs, \ba)$, we assume that the advantage values are random variables. We denote the advantage random variable for an action $\ba$ at a state $\bs$ with respect to policy $\pi$ with $A_\pi(\bs, \ba)$.

In the case of policies conditioned on advantages, the expected improvement of a policy $\pi(\ba|\bs, A)$ over a sampling policy $\mu(\ba|\bs, A)$ is given by 
\begin{equation}
    \eta_\mu(\pi) = \mathbb{E}_{\bs \sim d^\pi(\bs), {A} \sim p_\pi(A|\bs), \ba \sim \pi(\ba|\bs, {A})} \left[A_\mu(\bs, \ba) \right]
    \label{eqn:expec_improvement}
\end{equation}
When the policies $\mu$ and $\pi$ are close to each other, we obtain a trainable objective, by replacing the intractable state-distribution term $d^\pi(\bs)$ in Equation~\ref{eqn:expec_improvement} with state distribution $d^\mu(\bs)$ of the sampling policy. This approximation has been previously used in the derivation of TRPO~\cite{TRPOschulman15} and AWR~\cite{peng19awr}. For a rigorous proof of this approximation, we refer the readers to Lemma 3 in \citet{TRPOschulman15}. 

Our goal is to learn an advantage-conditioned policy $\pi(\ba|\bs, {A})$ which maximizes expected improvement while being close to the sampling policy $\mu(\ba|\bs, {A})$ while staying close to $\mu$ in distribution. This is formalized as the following optimization problem:
\begin{align}
    \mathop{\mathrm{arg \ max}}_{\pi} \quad &
    \mathbb{E}_{\bs \sim d^\mu(\bs), \ba, {A} \sim p_\mu(\ba, A|\bs)} \left[{A} \right]\\
    \textrm{s.t.} \quad & \mathbb{E}_{\bs \sim d^\mu(\bs)} \left[\mathrm{D_{KL}} \left( p_\pi(\ba, A|\bs) || p_\mu(\ba, A|\bs) \right) \right] \leq \varepsilon 
    \label{eqn:RCPCons0}
\end{align}
Following steps similar to the previous derivation for the return-conditioned variant (RCP-R), we obtain the following maximum-likelihood objective to train a parametric policy $\pi_\theta(a|s, \hat{A})$, given a sampling policy $\mu$, as described in Step 9 of Algorithm~\ref{alg:rcp}. 
\begin{equation}
    \pi_\theta(\ba|\bs, \hat{A}) = \mathop{\mathrm{arg \ max}_{\theta}} ~~ \mathbb{E}_{\bs \sim d^\mu(\bs), {A} \sim p_\mu(A|\bs)} \left[ \mathbb{E}_{\ba \sim \mu(\ba|\bs, {A})} \left[ \log \pi_\theta(\ba|\bs, {A}) \right] \right] 
    \label{eqn:acp_policy_update}
\end{equation}
Further, the target distribution of advantages at any state $\bs$ under this procedure is given by:
\begin{equation}
    p_{\pi^*}({A}|\bs) \propto p_\mu({A}|\bs) \exp\left(\frac{{A}}{\beta}\right)
    \label{eqn:adv_in_acp}
\end{equation}
To summarize, this derivation motivates a maximum-likelihood objective (Equation~\ref{eqn:acp_policy_update}) that trains the policy to choose actions that achieve a particular target advantage value as depicted in Algorithm~\ref{alg:rcp}, and the target distribution $p_{\pi^*}({A}|\bs)$ is updated according to Equation~\ref{eqn:adv_in_acp} to assign higher likelihoods to actions with higher advantages. Rather than fitting a model to learn a mapping between states and advantages, our model for the target distribution $p_{\pi^*}({A})$, as described in Section~\ref{sec:details} ignores the dependency on states in the interest of simplicity.  

\subsection{Weighted Maximum Likelihood for Reward-Conditioned Policy Learning}
\label{sec:weighting}

The derivation in Sections~\ref{sec:theory} gives rise to a simple maximum likelihood objective for training the reward-conditioned policy $\pi_\theta(\ba|\bs, Z)$. In contrast to prior work, such as REPS~\cite{Peters2010REP} and AWR~\cite{peng19awr}, which use a \emph{return-weighted} maximum likelihood objective to train an \emph{unconditioned} policy, with weights given by exponentiated returns, we expect our \emph{unweighted} maximum-likelihood objective to exhibit less variance, since exponentiated return weights necessarily reduce the effective sample size when many of the (suboptimal) trajectories receive very small weights. However, we can choose to also use weighted likelihood objective with RCPs, and indeed are free to prioritize the samples in $\mathcal{D}$ to attain better performance. For example, in the case of RCP-A, we can choose to upweight transitions corresponding to highly advantageous actions, rather than training under the data distribution defined by $\mathcal{D}$. As we show empirically in Section~\ref{sec:benchmarks}, prioritizing transition samples by assigning a weight proportional to exponential target value (either advantage or return) increases sample-efficiency in some cases, although this step is optional in the RCP framework. In practice, we would expect this to also reduce the effective sample size, though we did not find that to be a problem for the benchmark tasks on which we evaluated our method.

\section{Experimental Evaluation}

Our experiments aim to evaluate the performance of RCPs on standard RL benchmark tasks, as well as fully off-policy RL problems. We also present an ablation analysis, which aims to answer the following questions: \textbf{(1)} Do RCPs actually achieve a return that matches the value they are conditioned on? \textbf{(2)} What is the effect of the policy architecture on the performance of RCPs? \textbf{(3)} How does the choice of reweighting method during supervised learning affect performance, and can RCPs perform well with no reweighting at all? \textbf{(4)} Are RCPs less sensitive to the size of the replay buffer, as compared to other RL algorithms that use supervised subroutines, such as AWR? 

\paragraph{Experimental setup.} At each iteration, RCP collects 2000 transition samples (i.e. executes 2000 timesteps in the environment),
which are appended to the dataset $\mathcal{D}$. Unless stated otherwise, for RCPs, $\mathcal{D}$ is a ring buffer that holds 100k transitions. We also show results with larger buffer sizes in Figure~\ref{fig:buffer_size_abation}. Updates to the policy are performed by uniformly sampling minibatches of 256 samples from $\mathcal{D}$. For the advantage-conditioned variant, the value function is updated with 200 gradient steps per iteration, and the policy is updated with 1000 steps.

\subsection{Performance and Comparisons on Standard Benchmarks}
\label{sec:benchmarks}

We compare RCP-R and RCP-A to a number of prior RL algorithms, including on-policy algorithms such as TRPO~\cite{TRPOschulman15} and PPO~\cite{PPOSchulmanWDRK17}, and off-policy algorithms such as SAC~\cite{haarnoja18b} and DDPG~\cite{DDPG2016}. We also compare to AWR~\cite{peng19awr}, a recently proposed off-policy RL method that also utilizes supervised learning as a subroutine, but does not condition on rewards and requires an exponential weighting scheme during training. When using exponential weighting, both RCP-R and RCP-A resemble AWR, with the main difference being the additional conditioning on returns or advantages. However, RCPs can also use unweighted supervised learning, which can decrease the variance of the supervised learning stage and increase the effective sample size, while AWR requires exponential weighting, without which it can never learn an optimal policy.

\begin{figure}[t]
	\centering
    {\includegraphics[width=0.24\linewidth]{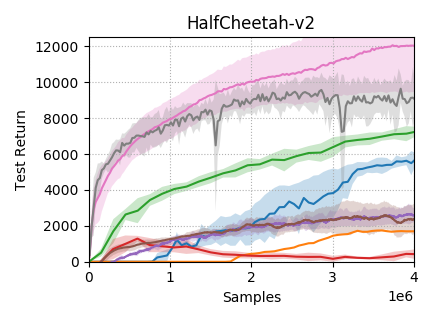}}
    {\includegraphics[width=0.24\linewidth]{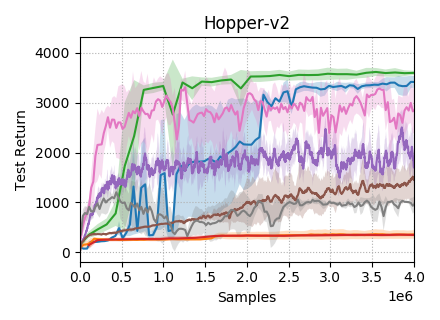}}
    {\includegraphics[width=0.24\linewidth]{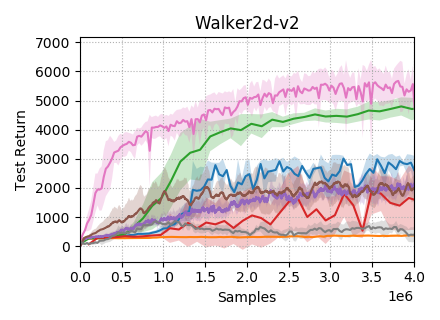}}
    {\includegraphics[width=0.24\linewidth]{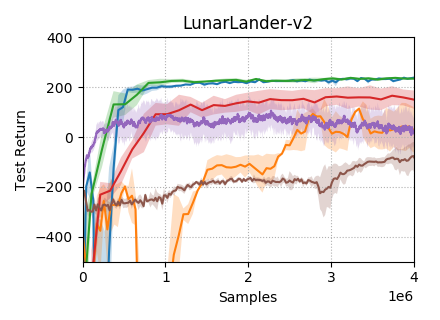}}\\
    {\includegraphics[width=0.8\columnwidth]{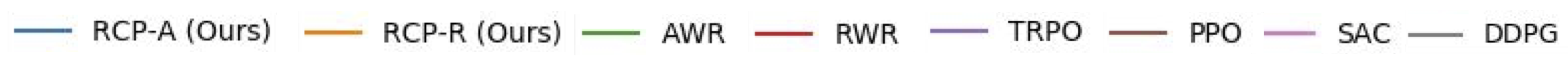}}
    \vspace{-0.25cm}
    \caption{\footnotesize{Learning curves of the various algorithms when applied to benchmark tasks. Results are averaged across 5 random seeds. RCP-R performs at par with RWR, and RCP-A is able to learn successful policies for each of the tasks, often outperforming several prior methods.}}
\label{fig:learningCurvesGym}
\end{figure}

\begin{figure}[t]
	\centering
    {\includegraphics[width=0.24\linewidth]{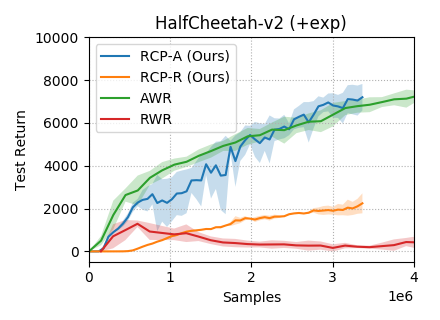}}
    {\includegraphics[width=0.24\linewidth]{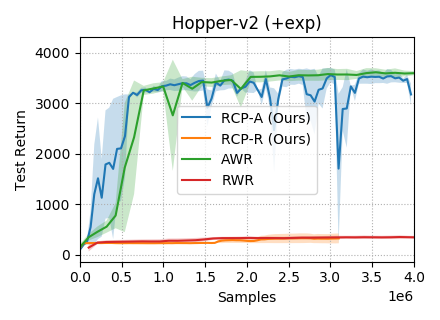}}
    {\includegraphics[width=0.24\linewidth]{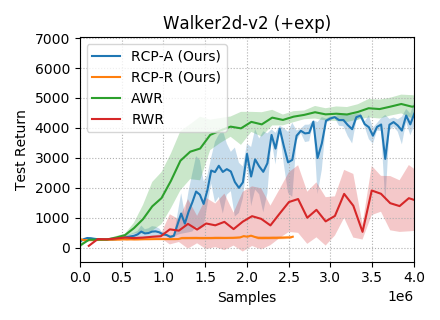}}
    {\includegraphics[width=0.24\linewidth]{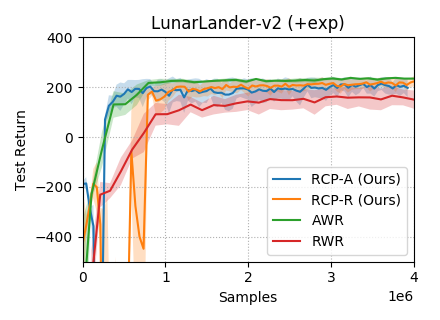}}
\vspace{-0.25cm}
\caption{\footnotesize{Learning curves for RCP-A and RCP-R with exponential weights for training the policy. AWR is shown for comparison. Results are averaged across 5 random seeds. RCP-A performs similarly to AWR when exponential weighting is used.}}
\label{fig:learningCurvesGymwithExp}
\end{figure}

Learning curves comparing the different algorithms on three continuous-control and one discrete-action OpenAI gym benchmark tasks are shown in Figure~\ref{fig:learningCurvesGym}. RCP-A substantially outperforms the return-conditioned variant, RCP-R, on all of the tasks, though RCP-R is still able to learn effective policies on the LundarLander-v2 task. While there is still a gap between the performance of RCPs and the best current reinforcement learning algorithms, RCP-A outperforms TRPO and performs comparably or better to PPO. When we additionally incorporate exponential reweighting, as shown in Figure~\ref{fig:learningCurvesGymwithExp}, both variants of RCP perform substantially better, and RCP-A performs similarly to AWR, though this is in a sense not surprising, since both methods utilize the same weighted regression step, with the only difference being that the RCP-A policy also receives the advantage values as an input. These results show that, although there is still a gap in performance between RCPs and prior methods, the methods has the potential to learn effective policies on a range of benchmark tasks.

As noted in Section~\ref{sec:related}, concurrently to our work, \citet{schmidhuber2019upsidedownrl} proposed a similar approach, UDRL, though without weighting or advantage conditioning, and reports a final result of around $150$ on the LunarLander-v2 task. We can see in Figure~\ref{fig:learningCurvesGym} that RCPs generally perform better, with RCP-A reaching $238 \pm 1.3$ on the same task. This suggests that, although conditioning on rewards provides for a simple and effective reinforcement learning method, there are still a number of simple but important design decisions that are essential for good performance.

\subsection{Performance in Fully Offline Settings}
\label{sec:batch_rl}
Since RCPs use standard supervised learning and can utilize all previously collected data, we would expect RCPs to be well suited for learning entirely from offline datasets, without on-policy data collection. We follow the protocol described by \citet{BEAR2019} and evaluate on static datasets collected from a ``mediocre'' partially trained policy, with 1 million transition samples per task.
RCPs can be trained directly on this dataset, without any modification to the algorithm. We compare to
AWR~\cite{peng19awr} and bootstrapping error accumulation reduction (BEAR)~\cite{BEAR2019}, which is a Q-learning method that incorporates a constraint to handle out-of-distribution actions. We also compare to off-policy approximate dynamic programming methods primarily designed for online learning -- SAC~\cite{haarnoja18b} and TD3~\cite{fujimoto18a}, -- and PPO~\cite{PPOSchulmanWDRK17}, which is an importance-sampled policy gradient algorithm.

\begin{wrapfigure}{r}{0.57\textwidth} 
    \centering
    \vspace{-0.1in}
    \includegraphics[width=0.47\linewidth]{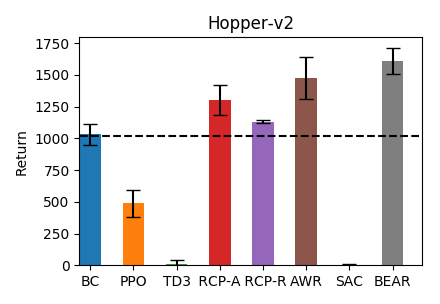}
    \includegraphics[width=0.47\linewidth]{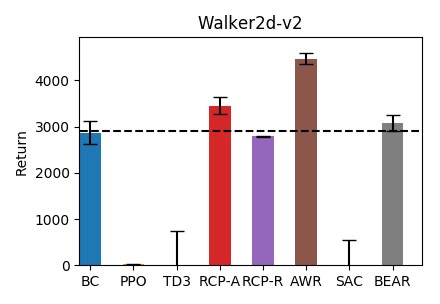}
    \caption{\footnotesize{Performance of various algorithms on fully off-policy learning tasks with static datasets. RCP-A learns effective policies that achieve better than dataset average in both cases. RCP-R performs similarly to behavioral cloning (BC).}}
    \vspace{-0.5in}
    \label{fig:batch_rl}
\end{wrapfigure}
As shown in Figure~\ref{fig:batch_rl}, we find that RCP-A learns effective policies in the purely offline setting on both the environments tested on and achieves performance better than the behavior policy that generated the dataset.

\subsection{Ablation Experiments}
\label{sec:ablations}

\begin{figure}
    \centering
    \begin{subfigure}{0.22\linewidth}
        \includegraphics[width=0.99\linewidth]{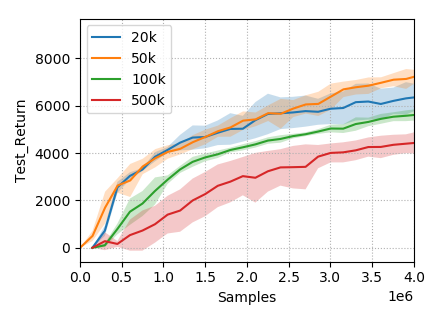}
        \caption{\footnotesize{AWR}}
    \end{subfigure}
    ~
    \begin{subfigure}{0.22\linewidth}
        \includegraphics[width=0.99\linewidth]{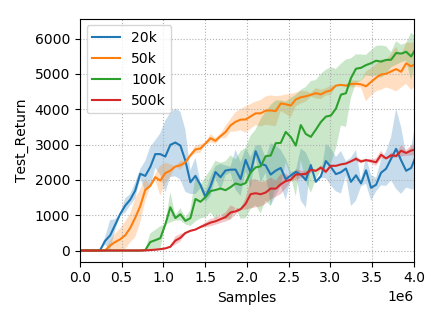}
        \caption{\footnotesize{RCP-A}}
    \end{subfigure}
    ~
    \begin{subfigure}{0.22\linewidth}
        \includegraphics[width=0.99\linewidth]{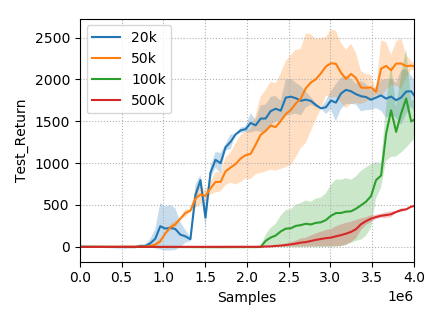}
        \caption{\footnotesize{RCP-R}}
    \end{subfigure}
    ~
     \begin{subfigure}{0.23\linewidth}
        \includegraphics[width=0.99\linewidth]{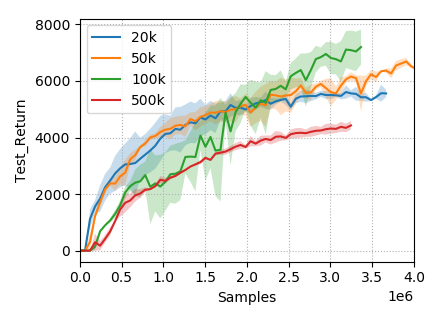}
        \caption{\footnotesize{RCP-A (+exp)}}
    \end{subfigure}
    \caption{\footnotesize{Learning curves demonstrating the effect of varying buffer sizes (20k, 50k, 100k and 500k) on different algorithms: (a) AWR (b) RCP-A (c) RCP-R and (d) RCP-A with exponential weighting on the HalfCheetah-v2 benchmark task. RCP-A generally performs better with larger buffers (compare 50k vs 100k), though performance still degrades with larger buffers.}}
    \label{fig:buffer_size_abation}
\end{figure}

Finally, we perform three ablation experiments to determine the effect of various design decisions for RCP training. The first parameter of variation is the size of the buffer $\mathcal{D}$ that is used during training. We compare RCP-R, RCP-A, and AWR with different buffer sizes, shown in Figure~\ref{fig:buffer_size_abation}. Note that the performance of AWR degrades significantly as the buffer size increases. This is expected, since AWR constrains the policy against the buffer distribution, therefore larger buffer sizes can result in slower policy improvement. In contrast, RCP-R and RCP-A can handle larger buffers, and perform better with buffers of size 100k as compared to buffers of size 50k, though larger buffers still result in somewhat worse performance. We speculate that this might be due to the fact the low-dimensional and simple benchmark task do not actually require large datasets to train an effective policy, and we might expect larger buffers to be more beneficial on more complex tasks, which we hope to investigate in the future.

\begin{figure}
    \centering
    \begin{subfigure}{0.23\linewidth}
        \centering
        \includegraphics[width=0.99\linewidth]{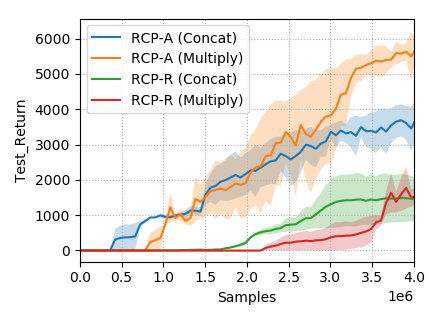}
        \caption{\centering \footnotesize{HalfCheetah-v2\break (RCP-A/ RCP-R)}}
    \end{subfigure}
    ~
    \begin{subfigure}{0.23\linewidth}
        \centering
        \includegraphics[width=0.99\linewidth]{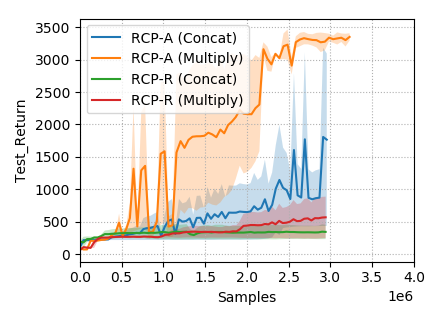}
        \caption{\centering \footnotesize{Hopper-v2\break (RCP-A/ RCP-R)}}
    \end{subfigure}
    ~
    \begin{subfigure}{0.23\linewidth}
        \centering
        \includegraphics[width=0.99\linewidth]{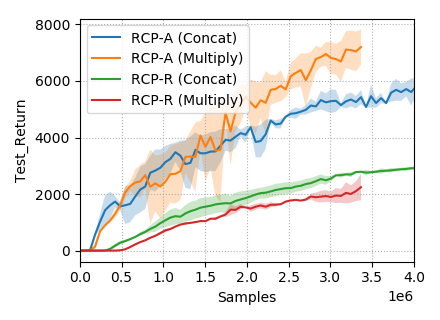}
        \caption{\centering \footnotesize{HalfCheetah-v2\break (+ exp weights)}}
    \end{subfigure}
    ~
     \begin{subfigure}{0.23\linewidth}
        \centering
        \includegraphics[width=0.99\linewidth]{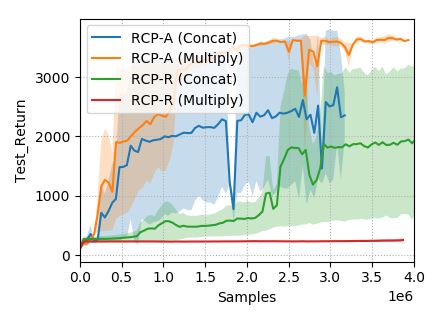}
        \caption{\centering \footnotesize{Hopper-v2\break (+ exp weights)}}
    \end{subfigure}
    \caption{\footnotesize{Performance of different architectures on HalfCheetah-v2 and Hopper-v2 environments with replay buffers of size 100k. Figures (c) and (d) correspond to weighted versions of both RCP-A and RCP-R. Note that the architecture \textit{multiply} clearly outperforms \textit{concat} in all cases.}}
    \label{fig:arch_ablation}
\end{figure}

In Figure~\ref{fig:arch_ablation}, we compare two different architectural choices for both the RCP variants. In the first architecture, labeled \emph{concat} in Figure~\ref{fig:arch_ablation}, the target value $Z$ is concatenated to the state $\mathbf{s}$ and then fed into a three-layer fully-connected network. The second architecture, labeled as \emph{multiply}, is our default choice for experiments in Section~\ref{sec:benchmarks} and uses multiplicative interactions, as discussed in Section~\ref{sec:details}. Learning curves in Figure~\ref{fig:arch_ablation} show that the architecture with multiplicative interactions (multiply) leads to better performance across the different environments (HalfCheetah-v2 and Hopper-v2) for both variants (RCP-A and RCP-R).

\begin{figure}
    \centering
   \begin{subfigure}{0.47\linewidth}
        \centering
        \includegraphics[width=0.99\linewidth]{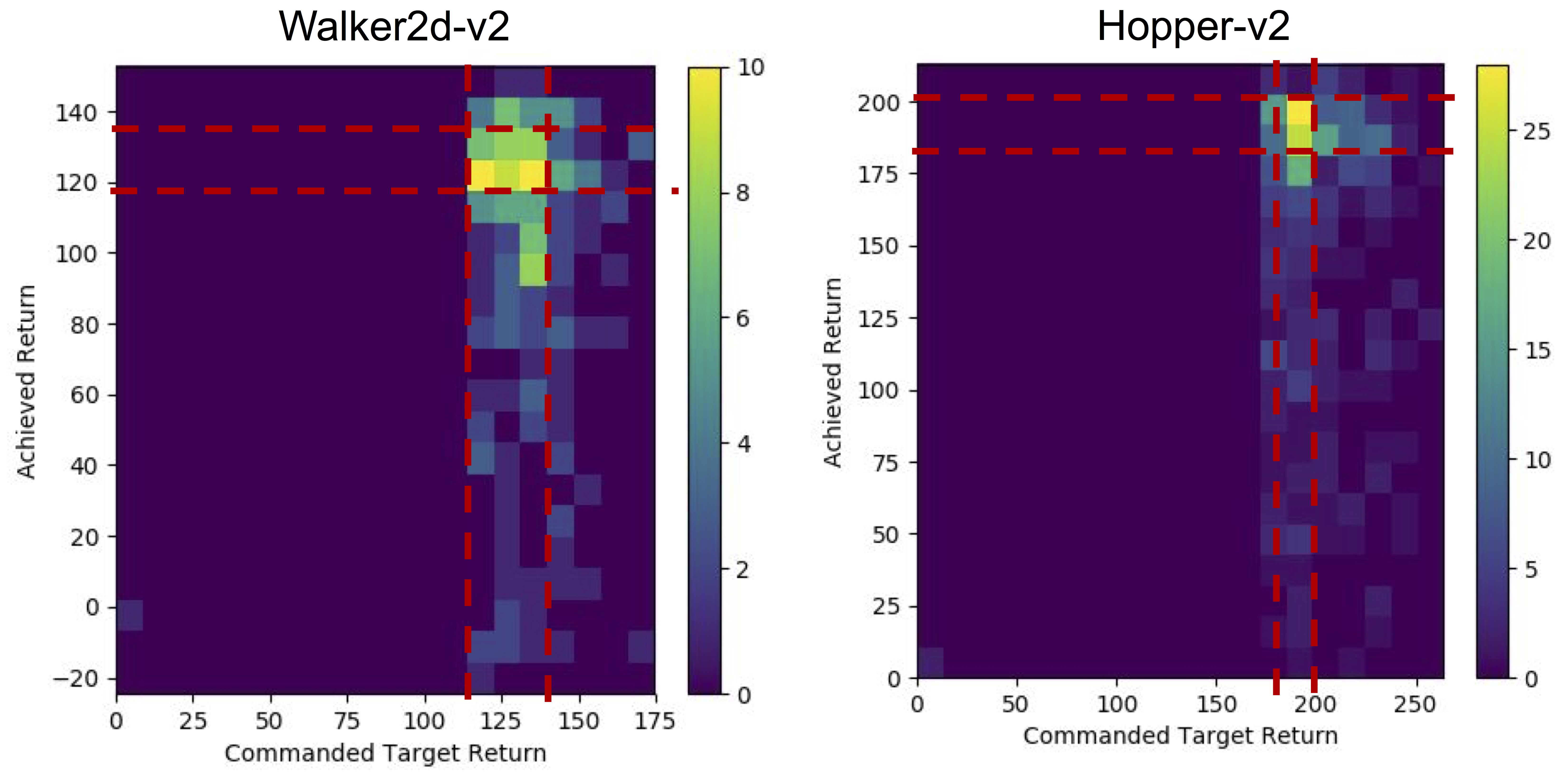}
        \caption{\centering \footnotesize{Target vs observed trajectory return (RCP-R)}}
    \end{subfigure}
    ~
    \begin{subfigure}{0.50\linewidth}
        \centering
        \includegraphics[width=0.96\linewidth]{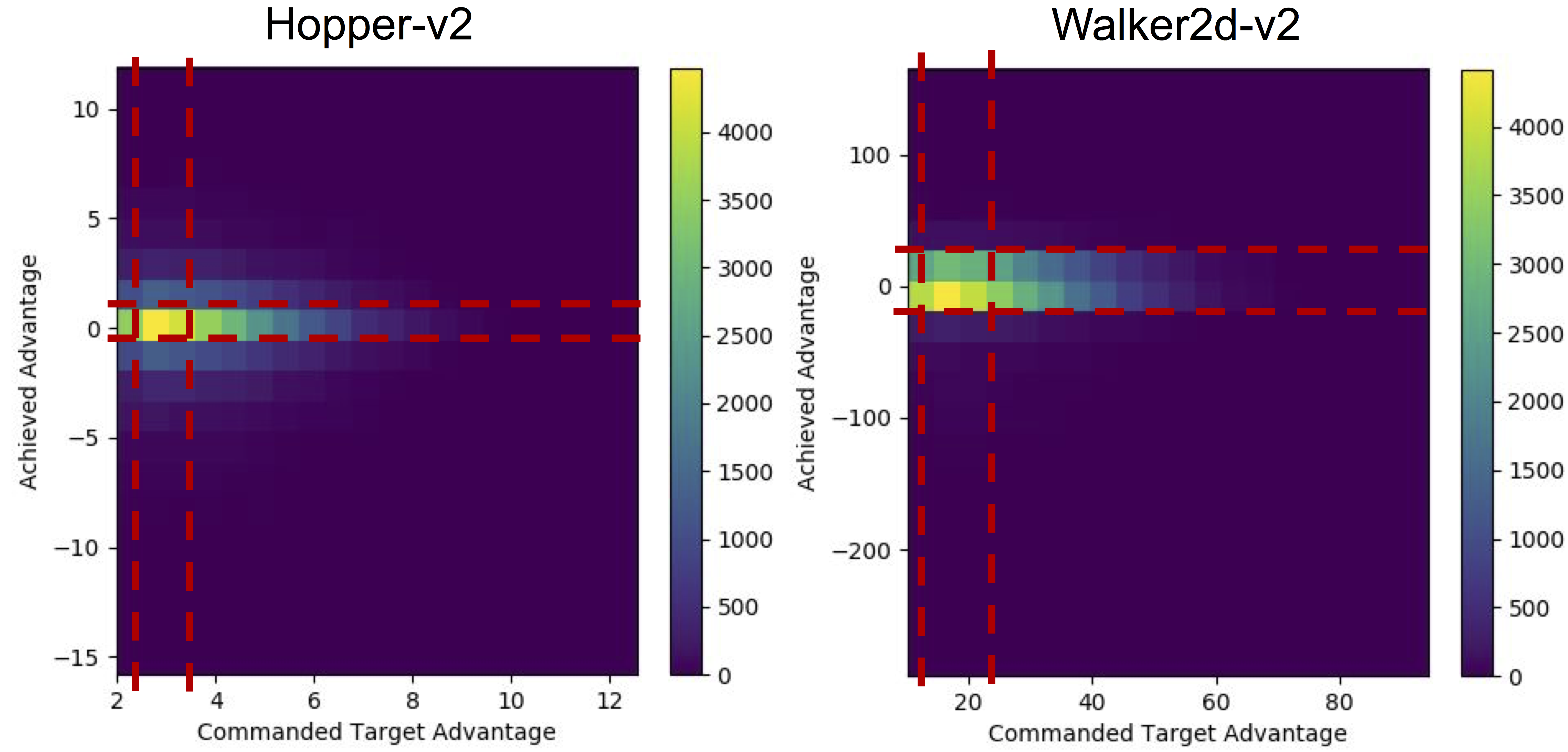}
        \caption{\centering \footnotesize{Target vs obtained action advantages (RCP-A)}}
    \end{subfigure}
    ~
    \caption{\footnotesize{Two-dimensional heatmap visualizing the co-occurrence frequencies of the specified target value $\hat{Z}$ (x-axis) and the observed value ${Z}$ (y-axis) after 2000 epochs of training for (a) RCP-R and (b) RCP-A. The co-occurrence frequencies are empirically estimated using separately executed rollouts that are conditioned on target values sampled from the instantaneous target model $p_{\pi^*}(Z)$. Note the similar magnitudes of $Z$ and $\hat{Z}$ in most cases.}} 
    \label{fig:commmanded_vs_obtained}
\end{figure}

Finally, we study the relationship between the target value $\hat{Z}$ that the policy is conditioned and the observed target value $Z$ achieved by rollouts from the policy. Ideally, we would expect the specified target values of $Z$ to roughly match the observed value $\hat{Z}$, as a reward-conditioned policy is explicitly trained to ensure this (Step 9 of Algorithm 1). In this experiment, we plot a two-dimensional heatmap of co-occurrence frequencies of $\hat{Z}$ and $Z_t$ to visualize the relationship between these quantities after about 2000 training iterations for both RCP variants. These heatmaps are shown in Figure~\ref{fig:commmanded_vs_obtained}. We find that both variants of RCP policies achieve returns (or advantages) that are similar enough to the target values they are conditioned on. 

\section{Discussion and Future Work}

We presented reward-conditioned policies, a general class of algorithms that enable learning of control policies with standard supervised learning approaches. Reward-conditioned policies make use of a simple idea: sub-optimal trajectories can be regarded as optimal supervision for a policy that does not aim to attain the largest possible reward, but rather to match the reward of that trajectory. By then conditioning the policy on the reward, we can train a single model to simultaneously represent policies for all possible reward values, and generalize to larger reward values.

While our results demonstrate that this approach can attain good results across a range of reinforcement learning benchmark tasks, its sample efficiency and final performance still lags behind the best and most efficient approximate dynamic programming methods, such as soft actor-critic~\citep{haarnoja18b}, as well as methods that utilize supervised learning in concert with reweighting, such as AWR~\citep{peng19awr}.
sWe nonetheless expect the simplicity of RCPs to serve as a significant benefit in many practical situations, and we hope that the use of standard supervised learning as a subroutine can also make it easier to analyze and understand the propoerties of our method. We expect that exploration is likely to be one of the major challenges with reward-conditioned policies: the methods we presented rely on generalization and random chance to acquire trajectories that improve in performance over those previously seen in the dataset. Sometimes the reward-conditioned policies might generalize successfully, and sometimes they might not. Further theoretical and empirical analysis of this generalization behavior may lead to a more performant class of methods, and more optimal sampling strategies inspired by posterior sampling may also lead to better final results. We believe that investigating these directions is an exciting avenue for future work, as it might allow us to devise a new class of reinforcement learning methods that combine the ease of use of supervised learning with the ability to autonomously acquire near-optimal behaviors from only high-level reward specification.

\bibliographystyle{plainnat}
\bibliography{main}

\begin{thebibliography}{48}
\providecommand{\natexlab}[1]{#1}
\providecommand{\url}[1]{\texttt{#1}}
\expandafter\ifx\csname urlstyle\endcsname\relax
  \providecommand{\doi}[1]{doi: #1}\else
  \providecommand{\doi}{doi: \begingroup \urlstyle{rm}\Url}\fi

\bibitem[Andrychowicz et~al.(2017)Andrychowicz, Wolski, Ray, Schneider, Fong,
  Welinder, McGrew, Tobin, Abbeel, and Zaremba]{Andrychowicz2017her}
Marcin Andrychowicz, Filip Wolski, Alex Ray, Jonas Schneider, Rachel Fong,
  Peter Welinder, Bob McGrew, Josh Tobin, Pieter Abbeel, and Wojciech Zaremba.
\newblock Hindsight experience replay.
\newblock In \emph{Proceedings of the 31st International Conference on Neural
  Information Processing Systems}, NIPS'17, pages 5055--5065, USA, 2017. Curran
  Associates Inc.
\newblock ISBN 978-1-5108-6096-4.
\newblock URL \url{http://dl.acm.org/citation.cfm?id=3295222.3295258}.

\bibitem[{Balaguer} and {Carpin}(2011)]{combining_imitation_and_rl}
B.~{Balaguer} and S.~{Carpin}.
\newblock Combining imitation and reinforcement learning to fold deformable
  planar objects.
\newblock In \emph{2011 IEEE/RSJ International Conference on Intelligent Robots
  and Systems}, pages 1405--1412, Sep. 2011.
\newblock \doi{10.1109/IROS.2011.6094992}.

\bibitem[Codevilla et~al.(2017)Codevilla, M{\"{u}}ller, Dosovitskiy,
  L{\'{o}}pez, and Koltun]{imitation_endtoend}
Felipe Codevilla, Matthias M{\"{u}}ller, Alexey Dosovitskiy, Antonio
  L{\'{o}}pez, and Vladlen Koltun.
\newblock End-to-end driving via conditional imitation learning.
\newblock \emph{CoRR}, abs/1710.02410, 2017.
\newblock URL \url{http://arxiv.org/abs/1710.02410}.

\bibitem[de~Vries et~al.(2017)de~Vries, Strub, Mary, Larochelle, Pietquin, and
  Courville]{cbn}
Harm de~Vries, Florian Strub, Jeremie Mary, Hugo Larochelle, Olivier Pietquin,
  and Aaron~C Courville.
\newblock Modulating early visual processing by language.
\newblock In \emph{NIPS}. 2017.

\bibitem[Fu et~al.(2019)Fu, Kumar, Soh, and Levine]{fu19diagnosing}
Justin Fu, Aviral Kumar, Matthew Soh, and Sergey Levine.
\newblock Diagnosing bottlenecks in deep q-learning algorithms.
\newblock In Kamalika Chaudhuri and Ruslan Salakhutdinov, editors,
  \emph{Proceedings of the 36th International Conference on Machine Learning},
  volume~97 of \emph{Proceedings of Machine Learning Research}, pages
  2021--2030, Long Beach, California, USA, 09--15 Jun 2019. PMLR.
\newblock URL \url{http://proceedings.mlr.press/v97/fu19a.html}.

\bibitem[Fujimoto et~al.(2018)Fujimoto, van Hoof, and Meger]{fujimoto18a}
Scott Fujimoto, Herke van Hoof, and David Meger.
\newblock Addressing function approximation error in actor-critic methods.
\newblock In Jennifer Dy and Andreas Krause, editors, \emph{Proceedings of the
  35th International Conference on Machine Learning}, volume~80 of
  \emph{Proceedings of Machine Learning Research}, pages 1587--1596,
  Stockholmsmässan, Stockholm Sweden, 10--15 Jul 2018. PMLR.
\newblock URL \url{http://proceedings.mlr.press/v80/fujimoto18a.html}.

\bibitem[Ghosh et~al.(2018)Ghosh, Singh, Rajeswaran, Kumar, and Levine]{DnC}
Dibya Ghosh, Avi Singh, Aravind Rajeswaran, Vikash Kumar, and Sergey Levine.
\newblock Divide-and-conquer reinforcement learning.
\newblock In \emph{ICLR}, 2018.

\bibitem[{Ghosh} et~al.(2019){Ghosh}, {Gupta}, {Fu}, {Reddy}, {Devin},
  {Eysenbach}, and {Levine}]{ghosh2019gcsl}
Dibya {Ghosh}, Abhishek {Gupta}, Justin {Fu}, Ashwin {Reddy}, Coline {Devin},
  Benjamin {Eysenbach}, and Sergey {Levine}.
\newblock {Learning To Reach Goals Without Reinforcement Learning}.
\newblock \emph{arXiv e-prints}, art. arXiv:1912.06088, Dec 2019.

\bibitem[Gu et~al.(2016)Gu, Lillicrap, Sutskever, and Levine]{NAF16}
Shixiang Gu, Timothy Lillicrap, Ilya Sutskever, and Sergey Levine.
\newblock Continuous deep q-learning with model-based acceleration.
\newblock In Maria~Florina Balcan and Kilian~Q. Weinberger, editors,
  \emph{Proceedings of The 33rd International Conference on Machine Learning},
  volume~48 of \emph{Proceedings of Machine Learning Research}, pages
  2829--2838, New York, New York, USA, 20--22 Jun 2016. PMLR.
\newblock URL \url{http://proceedings.mlr.press/v48/gu16.html}.

\bibitem[Haarnoja et~al.(2018)Haarnoja, Zhou, Abbeel, and Levine]{haarnoja18b}
Tuomas Haarnoja, Aurick Zhou, Pieter Abbeel, and Sergey Levine.
\newblock Soft actor-critic: Off-policy maximum entropy deep reinforcement
  learning with a stochastic actor.
\newblock In Jennifer Dy and Andreas Krause, editors, \emph{Proceedings of the
  35th International Conference on Machine Learning}, volume~80 of
  \emph{Proceedings of Machine Learning Research}, pages 1861--1870,
  Stockholmsmässan, Stockholm Sweden, 10--15 Jul 2018. PMLR.
\newblock URL \url{http://proceedings.mlr.press/v80/haarnoja18b.html}.

\bibitem[{Harutyunyan} et~al.(2019){Harutyunyan}, {Dabney}, {Mesnard}, {Azar},
  {Piot}, {Heess}, {van Hasselt}, {Wayne}, {Singh}, {Precup}, and
  {Munos}]{harutyunyan2019hca}
Anna {Harutyunyan}, Will {Dabney}, Thomas {Mesnard}, Mohammad {Azar}, Bilal
  {Piot}, Nicolas {Heess}, Hado {van Hasselt}, Greg {Wayne}, Satinder {Singh},
  Doina {Precup}, and Remi {Munos}.
\newblock {Hindsight Credit Assignment}.
\newblock \emph{arXiv e-prints}, art. arXiv:1912.02503, Dec 2019.

\bibitem[Hasselt et~al.(2016)Hasselt, Guez, and Silver]{Hasselt2016}
Hado~van Hasselt, Arthur Guez, and David Silver.
\newblock Deep reinforcement learning with double q-learning.
\newblock In \emph{Proceedings of the Thirtieth AAAI Conference on Artificial
  Intelligence}, AAAI'16, pages 2094--2100. AAAI Press, 2016.
\newblock URL \url{http://dl.acm.org/citation.cfm?id=3016100.3016191}.

\bibitem[Heess et~al.(2017)Heess, TB, Sriram, Lemmon, Merel, Wayne, Tassa,
  Erez, Wang, Eslami, Riedmiller, and Silver]{HeessTSLMWTEWER17}
Nicolas Heess, Dhruva TB, Srinivasan Sriram, Jay Lemmon, Josh Merel, Greg
  Wayne, Yuval Tassa, Tom Erez, Ziyu Wang, S.~M.~Ali Eslami, Martin~A.
  Riedmiller, and David Silver.
\newblock Emergence of locomotion behaviours in rich environments.
\newblock \emph{CoRR}, abs/1707.02286, 2017.
\newblock URL \url{http://arxiv.org/abs/1707.02286}.

\bibitem[Henderson et~al.(2017)Henderson, Islam, Bachman, Pineau, Precup, and
  Meger]{henderson2017reinforcement}
Peter Henderson, Riashat Islam, Philip Bachman, Joelle Pineau, Doina Precup,
  and David Meger.
\newblock Deep reinforcement learning that matters, 2017.
\newblock URL \url{http://arxiv.org/abs/1709.06560}.
\newblock cite arxiv:1709.06560Comment: Accepted to the Thirthy-Second AAAI
  Conference On Artificial Intelligence (AAAI), 2018.

\bibitem[Hessel et~al.(2017)Hessel, Modayil, van Hasselt, Schaul, Ostrovski,
  Dabney, Horgan, Piot, Azar, and Silver]{RainbowDQN2017}
Matteo Hessel, Joseph Modayil, Hado van Hasselt, Tom Schaul, Georg Ostrovski,
  Will Dabney, Daniel Horgan, Bilal Piot, Mohammad~Gheshlaghi Azar, and David
  Silver.
\newblock Rainbow: Combining improvements in deep reinforcement learning.
\newblock \emph{CoRR}, abs/1710.02298, 2017.
\newblock URL \url{http://arxiv.org/abs/1710.02298}.

\bibitem[Kaelbling(1993)]{Kaelbling93b}
Leslie~Pack Kaelbling.
\newblock Learning to achieve goals.
\newblock In \emph{Proceedings of the Thirteenth International Joint Conference
  on Artificial Intelligence}, Chambery, France, 1993. Morgan Kaufmann.
\newblock URL \url{http://people.csail.mit.edu/lpk/papers/ijcai93.ps}.

\bibitem[Kakade and Langford(2002)]{Kakade2002}
Sham Kakade and John Langford.
\newblock Approximately optimal approximate reinforcement learning.
\newblock In \emph{Proceedings of the Nineteenth International Conference on
  Machine Learning}, ICML '02, pages 267--274, San Francisco, CA, USA, 2002.
  Morgan Kaufmann Publishers Inc.
\newblock ISBN 1-55860-873-7.
\newblock URL \url{http://dl.acm.org/citation.cfm?id=645531.656005}.

\bibitem[Kalashnikov et~al.(2018)Kalashnikov, Irpan, Pastor, Ibarz, Herzog,
  Jang, Quillen, Holly, Kalakrishnan, Vanhoucke, and
  Levine]{kalashnikov18qtopt}
Dmitry Kalashnikov, Alex Irpan, Peter Pastor, Julian Ibarz, Alexander Herzog,
  Eric Jang, Deirdre Quillen, Ethan Holly, Mrinal Kalakrishnan, Vincent
  Vanhoucke, and Sergey Levine.
\newblock Qt-opt: Scalable deep reinforcement learning for vision-based robotic
  manipulation.
\newblock \emph{CoRR}, abs/1806.10293, 2018.
\newblock URL \url{http://arxiv.org/abs/1806.10293}.

\bibitem[Kumar et~al.(2019)Kumar, Fu, Tucker, and Levine]{BEAR2019}
Aviral Kumar, Justin Fu, George Tucker, and Sergey Levine.
\newblock Stabilizing off-policy q-learning via bootstrapping error reduction.
\newblock \emph{CoRR}, abs/1906.00949, 2019.
\newblock URL \url{http://arxiv.org/abs/1906.00949}.

\bibitem[Levine et~al.(2016)Levine, Finn, Darrell, and Abbeel]{Levine2016}
Sergey Levine, Chelsea Finn, Trevor Darrell, and Pieter Abbeel.
\newblock End-to-end training of deep visuomotor policies.
\newblock \emph{J. Mach. Learn. Res.}, 17\penalty0 (1):\penalty0 1334--1373,
  January 2016.
\newblock ISSN 1532-4435.
\newblock URL \url{http://dl.acm.org/citation.cfm?id=2946645.2946684}.

\bibitem[Lillicrap et~al.(2016)Lillicrap, Hunt, Pritzel, Heess, Erez, Tassa,
  Silver, and Wierstra]{DDPG2016}
Timothy~P. Lillicrap, Jonathan~J. Hunt, Alexander Pritzel, Nicolas Manfred~Otto
  Heess, Tom Erez, Yuval Tassa, David Silver, and Daan Wierstra.
\newblock Continuous control with deep reinforcement learning.
\newblock \emph{ICLR}, 2016.

\bibitem[Mnih et~al.(2015)Mnih, Kavukcuoglu, Silver, Rusu, Veness, Bellemare,
  Graves, Riedmiller, Fidjeland, Ostrovski, Petersen, Beattie, Sadik,
  Antonoglou, King, Kumaran, Wierstra, Legg, and Hassabis]{mnih2015humanlevel}
Volodymyr Mnih, Koray Kavukcuoglu, David Silver, Andrei~A. Rusu, Joel Veness,
  Marc~G. Bellemare, Alex Graves, Martin Riedmiller, Andreas~K. Fidjeland,
  Georg Ostrovski, Stig Petersen, Charles Beattie, Amir Sadik, Ioannis
  Antonoglou, Helen King, Dharshan Kumaran, Daan Wierstra, Shane Legg, and
  Demis Hassabis.
\newblock Human-level control through deep reinforcement learning.
\newblock \emph{Nature}, 518\penalty0 (7540):\penalty0 529--533, February 2015.
\newblock ISSN 00280836.
\newblock URL \url{http://dx.doi.org/10.1038/nature14236}.

\bibitem[Munos et~al.(2016)Munos, Stepleton, Harutyunyan, and
  Bellemare]{Munos2016}
R{\'e}mi Munos, Thomas Stepleton, Anna Harutyunyan, and Marc~G. Bellemare.
\newblock Safe and efficient off-policy reinforcement learning.
\newblock In \emph{Proceedings of the 30th International Conference on Neural
  Information Processing Systems}, NIPS'16, pages 1054--1062, USA, 2016. Curran
  Associates Inc.
\newblock ISBN 978-1-5108-3881-9.
\newblock URL \url{http://dl.acm.org/citation.cfm?id=3157096.3157214}.

\bibitem[Nachum et~al.(2018)Nachum, Norouzi, Tucker, and
  Schuurmans]{nachum2018learning}
Ofir Nachum, Mohammad Norouzi, George Tucker, and Dale Schuurmans.
\newblock Learning gaussian policies from smoothed action value functions,
  2018.
\newblock URL \url{https://openreview.net/forum?id=B1nLkl-0Z}.

\bibitem[Oh et~al.(2018)Oh, Guo, Singh, and Lee]{SIL2018}
Junhyuk Oh, Yijie Guo, Satinder Singh, and Honglak Lee.
\newblock Self-imitation learning.
\newblock In Jennifer Dy and Andreas Krause, editors, \emph{Proceedings of the
  35th International Conference on Machine Learning}, volume~80 of
  \emph{Proceedings of Machine Learning Research}, pages 3878--3887,
  Stockholmsmässan, Stockholm Sweden, 10--15 Jul 2018. PMLR.
\newblock URL \url{http://proceedings.mlr.press/v80/oh18b.html}.

\bibitem[Oord et~al.(2016)Oord, Kalchbrenner, Vinyals, Espeholt, Graves, and
  Kavukcuoglu]{oord16pixelcnn}
A\"{a}ron van~den Oord, Nal Kalchbrenner, Oriol Vinyals, Lasse Espeholt, Alex
  Graves, and Koray Kavukcuoglu.
\newblock Conditional image generation with pixelcnn decoders.
\newblock In \emph{Proceedings of the 30th International Conference on Neural
  Information Processing Systems}, NIPS’16, page 4797–4805, Red Hook, NY,
  USA, 2016. Curran Associates Inc.
\newblock ISBN 9781510838819.

\bibitem[Osa et~al.(2018)Osa, Pajarinen, and Neumann]{imitation_tutorial}
Takayuki Osa, Joni Pajarinen, and Gerhard Neumann.
\newblock \emph{An Algorithmic Perspective on Imitation Learning}.
\newblock Now Publishers Inc., Hanover, MA, USA, 2018.
\newblock ISBN 168083410X.

\bibitem[Peng et~al.(2018)Peng, Abbeel, Levine, and van~de
  Panne]{2018-TOG-deepMimic}
Xue~Bin Peng, Pieter Abbeel, Sergey Levine, and Michiel van~de Panne.
\newblock Deepmimic: Example-guided deep reinforcement learning of
  physics-based character skills.
\newblock \emph{ACM Trans. Graph.}, 37\penalty0 (4):\penalty0 143:1--143:14,
  July 2018.
\newblock ISSN 0730-0301.
\newblock \doi{10.1145/3197517.3201311}.
\newblock URL \url{http://doi.acm.org/10.1145/3197517.3201311}.

\bibitem[{Peng} et~al.(2019){Peng}, {Kumar}, {Zhang}, and {Levine}]{peng19awr}
Xue~Bin {Peng}, Aviral {Kumar}, Grace {Zhang}, and Sergey {Levine}.
\newblock {Advantage-Weighted Regression: Simple and Scalable Off-Policy
  Reinforcement Learning}.
\newblock \emph{arXiv e-prints}, art. arXiv:1910.00177, Sep 2019.

\bibitem[Perez et~al.(2017)Perez, Strub, de~Vries, Dumoulin, and
  Courville]{film}
Ethan Perez, Florian Strub, Harm de~Vries, Vincent Dumoulin, and Aaron~C.
  Courville.
\newblock Film: Visual reasoning with a general conditioning layer.
\newblock In \emph{AAAI}, 2017.

\bibitem[Peters and Schaal(2007)]{Peters2007RWR}
Jan Peters and Stefan Schaal.
\newblock Reinforcement learning by reward-weighted regression for operational
  space control.
\newblock In \emph{Proceedings of the 24th International Conference on Machine
  Learning}, ICML '07, pages 745--750, New York, NY, USA, 2007. ACM.
\newblock ISBN 978-1-59593-793-3.
\newblock \doi{10.1145/1273496.1273590}.
\newblock URL \url{http://doi.acm.org/10.1145/1273496.1273590}.

\bibitem[Peters et~al.(2010)Peters, M\"{u}lling, and Alt\"{u}n]{Peters2010REP}
Jan Peters, Katharina M\"{u}lling, and Yasemin Alt\"{u}n.
\newblock Relative entropy policy search.
\newblock In \emph{Proceedings of the Twenty-Fourth AAAI Conference on
  Artificial Intelligence}, AAAI'10, pages 1607--1612. AAAI Press, 2010.
\newblock URL \url{http://dl.acm.org/citation.cfm?id=2898607.2898863}.

\bibitem[Pomerleau(1989)]{alvinn}
Dean~A. Pomerleau.
\newblock Advances in neural information processing systems 1.
\newblock chapter ALVINN: An Autonomous Land Vehicle in a Neural Network, pages
  305--313. Morgan Kaufmann Publishers Inc., San Francisco, CA, USA, 1989.
\newblock ISBN 1-558-60015-9.
\newblock URL \url{http://dl.acm.org/citation.cfm?id=89851.89891}.

\bibitem[Pong et~al.(2018)Pong, Gu, Dalal, and Levine]{pong18tdm}
Vitchyr Pong, Shixiang Gu, Murtaza Dalal, and Sergey Levine.
\newblock Temporal difference models: Model-free deep {RL} for model-based
  control.
\newblock \emph{ICLR}, abs/1802.09081, 2018.
\newblock URL \url{http://arxiv.org/abs/1802.09081}.

\bibitem[Precup et~al.(2001)Precup, Sutton, and Dasgupta]{Precup2001}
Doina Precup, Richard~S. Sutton, and Sanjoy Dasgupta.
\newblock Off-policy temporal difference learning with function approximation.
\newblock In \emph{Proceedings of the Eighteenth International Conference on
  Machine Learning}, ICML '01, pages 417--424, San Francisco, CA, USA, 2001.
  Morgan Kaufmann Publishers Inc.
\newblock ISBN 1-55860-778-1.
\newblock URL \url{http://dl.acm.org/citation.cfm?id=645530.655817}.

\bibitem[Rajeswaran et~al.(2018)Rajeswaran, Kumar, Gupta, Vezzani, Schulman,
  Todorov, and Levine]{Rajeswaran-RSS-18}
Aravind Rajeswaran, Vikash Kumar, Abhishek Gupta, Giulia Vezzani, John
  Schulman, Emanuel Todorov, and Sergey Levine.
\newblock {Learning Complex Dexterous Manipulation with Deep Reinforcement
  Learning and Demonstrations}.
\newblock In \emph{Proceedings of Robotics: Science and Systems (RSS)}, 2018.

\bibitem[{Schmidhuber}(2019)]{schmidhuber2019upsidedownrl}
Juergen {Schmidhuber}.
\newblock {Reinforcement Learning Upside Down: Don't Predict Rewards -- Just
  Map Them to Actions}.
\newblock \emph{arXiv e-prints}, art. arXiv:1912.02875, Dec 2019.

\bibitem[Schulman et~al.(2015)Schulman, Levine, Abbeel, Jordan, and
  Moritz]{TRPOschulman15}
John Schulman, Sergey Levine, Pieter Abbeel, Michael Jordan, and Philipp
  Moritz.
\newblock Trust region policy optimization.
\newblock In Francis Bach and David Blei, editors, \emph{Proceedings of the
  32nd International Conference on Machine Learning}, volume~37 of
  \emph{Proceedings of Machine Learning Research}, pages 1889--1897, Lille,
  France, 07--09 Jul 2015. PMLR.
\newblock URL \url{http://proceedings.mlr.press/v37/schulman15.html}.

\bibitem[Schulman et~al.(2016)Schulman, Moritz, Levine, Jordan, and
  Abbeel]{Schulmanetal_ICLR2016}
John Schulman, Philipp Moritz, Sergey Levine, Michael Jordan, and Pieter
  Abbeel.
\newblock High-dimensional continuous control using generalized advantage
  estimation.
\newblock In \emph{Proceedings of the International Conference on Learning
  Representations (ICLR)}, 2016.

\bibitem[Schulman et~al.(2017)Schulman, Wolski, Dhariwal, Radford, and
  Klimov]{PPOSchulmanWDRK17}
John Schulman, Filip Wolski, Prafulla Dhariwal, Alec Radford, and Oleg Klimov.
\newblock Proximal policy optimization algorithms.
\newblock \emph{CoRR}, abs/1707.06347, 2017.
\newblock URL \url{http://arxiv.org/abs/1707.06347}.

\bibitem[Singh et~al.(2019)Singh, Yang, Hartikainen, Finn, and
  Levine]{singh2019viceraq}
Avi Singh, Larry Yang, Kristian Hartikainen, Chelsea Finn, and Sergey Levine.
\newblock End-to-end robotic reinforcement learning without reward engineering.
\newblock \emph{CoRR}, abs/1904.07854, 2019.
\newblock URL \url{http://arxiv.org/abs/1904.07854}.

\bibitem[{Srivastava} et~al.(2019){Srivastava}, {Shyam}, {Mutz},
  {Ja{\'s}kowski}, and {Schmidhuber}]{srivastave2019upsidedownrl}
Rupesh~Kumar {Srivastava}, Pranav {Shyam}, Filipe {Mutz}, Wojciech
  {Ja{\'s}kowski}, and J{\"u}rgen {Schmidhuber}.
\newblock {Training Agents using Upside-Down Reinforcement Learning}.
\newblock \emph{arXiv e-prints}, art. arXiv:1912.02877, Dec 2019.

\bibitem[Sun et~al.(2018)Sun, Bagnell, and Boots]{sun2018truncated}
Wen Sun, J.~Andrew Bagnell, and Byron Boots.
\newblock {TRUNCATED} {HORIZON} {POLICY} {SEARCH}: {DEEP} {COMBINATION} {OF}
  {REINFORCEMENT} {AND} {IMITATION}.
\newblock In \emph{International Conference on Learning Representations}, 2018.
\newblock URL \url{https://openreview.net/forum?id=ryUlhzWCZ}.

\bibitem[Sutton et~al.(2000)Sutton, McAllester, Singh, and
  Mansour]{PoliGrad1999}
Richard~S Sutton, David~A. McAllester, Satinder~P. Singh, and Yishay Mansour.
\newblock Policy gradient methods for reinforcement learning with function
  approximation.
\newblock In S.~A. Solla, T.~K. Leen, and K.~M\"{u}ller, editors,
  \emph{Advances in Neural Information Processing Systems 12}, pages
  1057--1063. MIT Press, 2000.

\bibitem[Wang et~al.(2018)Wang, Xiong, Han, sun, Liu, and Zhang]{MARWIL}
Qing Wang, Jiechao Xiong, Lei Han, peng sun, Han Liu, and Tong Zhang.
\newblock Exponentially weighted imitation learning for batched historical
  data.
\newblock In S.~Bengio, H.~Wallach, H.~Larochelle, K.~Grauman, N.~Cesa-Bianchi,
  and R.~Garnett, editors, \emph{Advances in Neural Information Processing
  Systems 31}, pages 6288--6297. Curran Associates, Inc., 2018.

\bibitem[Wang et~al.(2016)Wang, Bapst, Heess, Mnih, Munos, Kavukcuoglu, and
  de~Freitas]{WangBHMMKF16}
Ziyu Wang, Victor Bapst, Nicolas Heess, Volodymyr Mnih, R{\'{e}}mi Munos, Koray
  Kavukcuoglu, and Nando de~Freitas.
\newblock Sample efficient actor-critic with experience replay.
\newblock \emph{CoRR}, abs/1611.01224, 2016.
\newblock URL \url{http://arxiv.org/abs/1611.01224}.

\bibitem[Watkins and Dayan(1992)]{Watkins92q}
Christopher J. C.~H. Watkins and Peter Dayan.
\newblock Q-learning.
\newblock In \emph{Machine Learning}, pages 279--292, 1992.

\bibitem[Williams(1992)]{Williams1992}
Ronald~J. Williams.
\newblock Simple statistical gradient-following algorithms for connectionist
  reinforcement learning.
\newblock \emph{Mach. Learn.}, 8\penalty0 (3-4):\penalty0 229--256, May 1992.
\newblock ISSN 0885-6125.
\newblock \doi{10.1007/BF00992696}.
\newblock URL \url{https://doi.org/10.1007/BF00992696}.

\end{thebibliography}

\end{document}